\documentclass[conference]{IEEEtran}
\IEEEoverridecommandlockouts
\usepackage[utf8]{inputenc}
\usepackage{cite}
\usepackage{amsthm,amsmath,amssymb,amsfonts}
\usepackage{algorithmic}
\usepackage{graphicx}
\usepackage{textcomp}
\usepackage{xcolor}
\usepackage{float}
\usepackage{tabularx}
\usepackage{enumitem}
\usepackage{makecell}
\usepackage{subcaption}
\usepackage{multirow}
\usepackage{algorithm}
\usepackage{algorithmic}
\usepackage{bbold}
\usepackage{array}
\usepackage{booktabs}
\usepackage{mathrsfs}
\usepackage[normalem]{ulem}
\usepackage{hyperref}
\usepackage{url}
\DeclareMathOperator*{\argmin}{arg\,min}

\def\BibTeX{{\rm B\kern-.05em{\sc i\kern-.025em b}\kern-.08em
    T\kern-.1667em\lower.7ex\hbox{E}\kern-.125emX}}

\begin{document}

\title{CeBed: A Benchmark for Deep Data-Driven \\ OFDM Channel Estimation\\
{}
}

\author{
    \IEEEauthorblockN{Amal Feriani, Di Wu, Xue Liu, Greg Dudek}
    \IEEEauthorblockA{Samsung AI Center Montreal, Canada
    \\\{amal.feriani, di.wu1, steve.liu, greg.dudek\}@samsung.com}
    }
\maketitle

\begin{abstract}
Deep learning has been extensively adopted in channel estimation problems. Although several data-driven approaches exist, a fair and reliable comparison between them has not been established yet due to the lack of a standardized experimental design. Data-driven approaches are often compared based on empirical results, thus the availability of standardized and reproducible evaluation tools is crucial for the development of such approaches.
In this work, we introduce the first Channel estimation testBed (CeBed) to unify several deep OFDM estimators.
First, we define a new taxonomy of scenarios covering various wireless system configurations and select a set of scenarios to standardize the training and evaluation conditions. Second, we perform a multi-metric (accuracy, robustness, generalization) analysis of the performance of ten deep and traditional baselines. Our evaluation provides insightful findings concerning the interplay between different metrics. For full transparency, we publicly release our implementation \footnote{https://github.com/SAIC-MONTREAL/CeBed}, offering a modular toolkit for easily adding new scenarios, models, and metrics.
 
\end{abstract}


\section{Introduction}
Channel estimation is one of the central problems in physical layer research to enable robust and reliable communications. Orthogonal frequency division multiplexing (OFDM) is the core transmission technology in 5G and beyond thanks to its high data-rate transmissions and resilience to frequency selective fading \cite{ofdm_mimo_survey}. Pilot-assisted channel estimation in OFDM systems relies on pilot symbols inserted in specific positions of the resource grid, known at the receiver and the transmitter \cite{cepilot}. 
Least square (LS) channel estimation uses the received signal and the transmitted pilots to estimate the channel impulse response by minimizing the mean squared error (MSE) between the received and estimated symbols \cite{ls}. Although the LS estimator has low complexity, it suffers from low estimation accuracy.
Linear minimum mean squared error (LMMSE) provides better accuracy at the expense of estimating the second-order statistics of the channel and the noise variance. However, the LMMSE method has a high complexity due to estimating the channel covariance matrix and is sensitive to estimation errors.

Deep learning (DL) has emerged as a powerful technique to solve complex problems in various fields including wireless communication. Several DL-based approaches have been proposed to alleviate the shortcomings of the traditional channel estimation methods and enhance the performance of the LS estimator. 
For instance, channel reconstruction methods \cite{channelNet, ReEsNet, InReEsNet, channelformer} use the LS estimated coefficients at pilot locations as low-resolution inputs and recover the whole channel matrix by learning an up-sampling and/or a denoising deep neural networks (DNNs). Another line of work relies on masked inputs with only coefficients at pilot positions known \cite{MTRE, DDAE} to estimate the channel coefficients at data positions. 

After carefully studying the aforementioned works, we observe two main concerns. First, the current literature lacks standardization and unification in both training and evaluation. Indeed, these DNNs are usually trained in a supervised manner using large datasets collected under different wireless environments in simulation. Thus, the evaluation of the trained models is highly dependent on the data collection process since a simple change in the environment's parameters can result in a distribution shift and hinder the performance of the DL models \cite{akrout2023domain}. Also, the experimental design (e.g., wireless system model, training and evaluation parameters) is often different across papers, thereby leading to scattered evaluation findings. Thus, it is important to define standardized conditions for training and evaluation to promote fair comparisons and identify failure points. The second concern is the lack of reproducible implementations since code and datasets are seldom released, except recent efforts \footnote{https://github.com/MehranSoltani94/ChannelNet}\footnote{https://github.com/dianixn/Channelformer}.
However, a careful investigation of these implementations reveals that they are either lacking the data generation process or do not include multiple deep baselines.

To address these concerns, we present the \emph{first} open benchmark for deep OFDM channel estimation, coined CeBed.  Specifically, we design standardized scenarios and evaluation protocols to rigorously and comprehensively compare deep channel estimators. CeBed introduces an abstract taxonomy of scenarios to define the design of the evaluation space. Our scenarios consist of different wireless system configurations spanning various system models, channel models, pilot designs and propagation environments. We implement a set of selected scenarios to conduct multi-metric analysis. We focus on three interconnected metrics: accuracy, generalization and robustness. Since it is not possible to cover the space of all scenarios, we performed targeted evaluation where for each metric, we define a standard set of scenarios used to train and evaluate all the baselines. To summarize, our contributions are:
\begin{itemize}[leftmargin=*]
    \item We introduce the Channel Estimation testBed (CeBed): a curated suite of scenarios, datasets and models for benchmarking the performance of deep channel estimators. We carefully implement \emph{ten} baselines (seven deep and three classical methods) and evaluate them on different scenarios spanning different channel models, system models, and propagation parameters;
    \item We release a unified framework to easily design reproducible and rigorous experimentation for new datasets and models. New methods and/or datasets can be easily integrated into CeBed in a few lines of code;
    \item We conduct extensive experiments with various DL-based models and analyze their performance based on different metrics. Our experiments yield insightful findings that can guide future model development or further analysis.
\end{itemize}
The rest of the paper is organized as follows. Section \ref{sec:background} reviews the channel estimation problem, traditional, and deep methods. The different components of CeBed are introduced in Section \ref{sec:cebed}. Next, we discuss the experimental results obtained by running CeBed. Finally, we conclude the work by discussing the lessons learned and sharing future research directions.
\section{Background}\label{sec:background}
\subsection{Channel Estimation Problem \& Traditional Methods}\label{sec:problem}
In OFDM systems \cite{ofdm}, the channel is a two-dimensional (2D) grid with $N_s$ symbols in time and $N_f$ sub-carriers in frequency. Considering a system with $N_r$ antennas at the receiver and a single antenna at the transmitter, the received signal in the frequency domain at \emph{each} receive antenna is:
\begin{align*}
    \mathbf{Y}_r = \mathbf{H}_r \odot \mathbf{X} + \mathbf{W}_r, \; r \in \{1,\dots, N_r\};
\end{align*}
where $\mathbf{X}$, $\mathbf{H}_r$, $\mathbf{Y}_r$ are $N_f \times N_s$ complex matrices denoting the transmitted symbols, the channel coefficients and received symbols of the $r$-th antenna, respectively. The symbol $\odot$ denotes the Hadamard product (element-wise product). The noise matrix $\mathbf{W}_r \in \mathbb{C}^{N_f\times N_s}$ contains an additive white Gaussian noise with zero mean and a variance $\sigma^2_w$.
Let $N_{fp}$ and $N_{sp}$ be the number of pilot sub-carriers and symbols, respectively. Thus, the channel estimation problem consists in estimating the channel coefficients $\mathbf{H}$ using the known transmitted pilots $\mathbf{X}_p \in \mathbb{C}^{N_{fp}\times N_{sp}}$ and the received pilot signals $\mathbf{Y}_p \in \mathbb{C}^{N_{fp}\times N_{sp}}$. 

The well-known LS method estimates the channel coefficients at pilot positions by dividing the received pilot symbols by the transmitted ones element-wise:
\begin{align*}
    \mathbf{H}_p^{LS} = \argmin_{H_p} || \mathbf{Y}_p - \mathbf{H}_p \odot \mathbf{X}_p ||_2^2 =  \frac{\mathbf{Y}_p}{\mathbf{X}_p}
\end{align*}
The channel coefficients at non-pilot positions are estimated using linear interpolation. 

The LMMSE approach improves upon the LS estimation by computing a filtering matrix $\mathbf{F}$ such that the error between the true channel and the LS estimates is minimized. The 2D LMMSE method uses the channel correlation from both time and frequency dimensions. The 2D LMMSE filtering can be separated into two one-dimensional steps: (1) frequency domain FD LMMSE and, (2) time-domain interpolation \cite{2dLMMSE}. The FD LMMSE estimates the channel coefficients \emph{only} for symbols that contain pilots as in: $\mathbf{F}_{i} =  \mathbf{R}_{\mathbf{h}_i,\mathbf{h}^i_p}\left( \mathbf{R}_{\mathbf{h}^i_p, \mathbf{h}^i_p} + \sigma_w^2 \mathbf{I}\right)^{-1}$
where $i$ is the index of the pilot symbol, $\mathbf{h}_i \in \mathbb{C}^{N_f}$ is the channel vector at the pilot symbol $i$ and $\mathbf{h}_p^i \in \mathbb{C}^{N_{fp}}$ is the vector of the channels
at the pilot locations in the frequency domain. $\mathbf{R}_{\mathbf{h}_i,\mathbf{h}_p^i} = \mathbb{E}\left[\mathbf{h}_i, \mathbf{h}_p^{i*}\right]$ and $\mathbf{R}_{\mathbf{h}_p^i,\mathbf{h}_p^i}=\mathbb{E}\left[\mathbf{h}_p^i\mathbf{h}_p^{i*}\right]$ are the cross-correlation matrix between $\mathbf{h}$ and $\mathbf{h}_p$ and the channel auto-correlation matrix at the pilot positions, respectively.
The time-domain interpolation applies any kind of interpolation to find the channel coefficients for all OFDM symbols.
To further reduce the complexity of the LMMSE method, approximate linear minimum mean square error (ALMMSE) is proposed \cite{ls}. It exploits the structure of the auto-correlation matrix for time and frequency correlations.
\subsection{Deep Learning Channel Estimation}
The channel estimation problem is similar to image restoration in digital image processing.

Specifically, the success of DL in super-resolution (SR) and image denoising inspired various deep channel estimation approaches \cite{channelNet, ReEsNet, InReEsNet}. These works consider the channel matrix at the pilot locations $\mathbf{H}_p^{LS}$ as low-resolution inputs and learn a DNN to reconstruct the whole channel $\mathbf{H}$. 
Current SR-based methods can be categorized into two families: \emph{pre-sampling} SR and \emph{post-sampling} SR. Pre-sampling SR methods first use traditional interpolation (e.g., bilinear or bicubic) to upsample the low-resolution inputs, afterwards, the interpolated inputs are refined using a DNN to obtain the final upsampled outputs. SR convolutional neural network (SRCNN) \cite{srcnn} is an example of a pre-sampling SR method used in previous art \cite{channelNet}. Pre-sampling SR methods have high complexity since the DNN is applied on the high-resolution space. To remedy this issue, post-sampling SR applies first the feature extraction on the low-resolution inputs and adds an upsampling layer at the end. The final upsampling layer can be either a traditional interpolation or a learned upsampling layer. Several channel estimation methods \cite{ReEsNet, InReEsNet} are based on a post-sampling SR technique called Enhanced Deep Residual Network (EDSR) where the SR network is built using a number of residual blocks \cite{edrsr}. The upsampled channel can be further denoised using a denoising network to further improve the estimation accuracy as in \cite{channelNet}.

An alternative formulation of the channel estimation problem is based on masked image modeling (MIM). In this case, the model receives a corrupt masked input where only a small portion is known and learns to predict the masked input signal from the unmasked portion. Since transformers \cite{transformer} have shown great success in this task, a transformer-based architecture is proposed to reconstruct the whole channel matrix from the masked LS inputs \cite{MTRE}. 
\begin{table*}[!ht]
    \centering
    \caption{Summary of the core scenarios.}
    \label{tab:task_summary}
    \begin{tabularx}{\textwidth}{X| l l l l l l |l l l l l l}
        Metric & \multicolumn{6}{c}{Train scenarios} & \multicolumn{6}{c}{Test scenarios} \\
        \cline{2-13}
         & $\mathbf{H}$ & $N_r$ & $N_{fp}$ & $N_{sp}$ & $\rho$ (dB) & $\upsilon$ (m/s) & $\mathbf{H}$ & $N_r$ & $N_{fp}$ & $N_{sp}$ & $\rho$ (dB) & $\upsilon$ (m/s) \\
        \hline
        \multirow{2}{10em}{Accuracy} & Umi & 1,4 & 72 & 2 & $[0,20]$ & $[0,15]$ & Umi & 1,4 & 72 & 2 & $[0,20]$ & $[0,15]$\\
        & Uma & 1,4 & 72 & 2 &  $[0,20]$ & $[0,15]$ & Uma & 1,4& 72 &2 &  $[0,20]$ & $[0,15]$ \\
        \hline
        \multirow{2}{10em}{Domain Generalization} & Umi & 1,4 & 72 &2 & $[0,20]$ & $[0,15]$ & Umi & 1,4 & 72 &2 & $[-30,0[\cup[25,30]$ & $[0,15]$\\
        & Uma & 1,4 & 72 &2 &  $[0,20]$ & $[0,15]$ & Uma & 1,4 & 72 & 2 &  $[-30,0[\cup[25,30]$ & $[0,15]$ \\
        \hline
        Robustness to Pilot Design & Umi & 1 & 72, 36 & 2,1 & $[0,20]$ & $15$ & Umi & 1 & 72,36 & 2,1 & $[0,20]$ & $15$\\
        \hline
        Robustness to Spatial Correlation & Umi & \thead{1,4\\8,16} & 72 &2 & $[0,20]$ & $5$ & Umi & \thead{1,4\\8,16} & 72 &2 & $[0,20]$ & $5$\\
        \bottomrule
    \end{tabularx}
\end{table*}
In CeBed, we implemented seven deep channel estimation baselines: four SR, and three MIM methods. These baselines are also based on different deep backbones (e.g., residual network, transformer). In the next section, we will first detail Cebed scenarios, afterwards we will describe the implemented baselines.
\addtolength{\topmargin}{0.03in}
\section{CEBED: A testbed for Channel Estimation}\label{sec:cebed}
In the literature, deep channel estimators are often trained and evaluated on specific scenarios used to generate the training and testing data. As an example, most of the prior work only considers a SISO system. Our objective is to build a holistic evaluation that requires training and evaluating the models under different wireless communication settings covering a wide space (e.g., channel models, noise levels, system models, etc). Next, we introduce an abstract taxonomy to help navigate this space.
\subsection{Taxonomy}\label{sec:cebed_datasets}
\noindent \textbf{Scenarios.} A scenario characterizes the wireless environment used to generate the training and/or test data. Given an OFDM wireless system, a scenario can be defined based on the key parameters that can impact the channel estimation problem which are the channel model, the system model, the pilot pattern, the user mobility and the noise level. Thus, a scenario is a $(\mathbf{H}, N_r, N_{fp}, N_{sp}, \rho, \upsilon)$ tuple where $\rho$ and $\upsilon$ denote the SNR level and the transmitter speed, respectively.
\begin{figure}[ht!]
    \centering
    \includegraphics[scale=0.39]{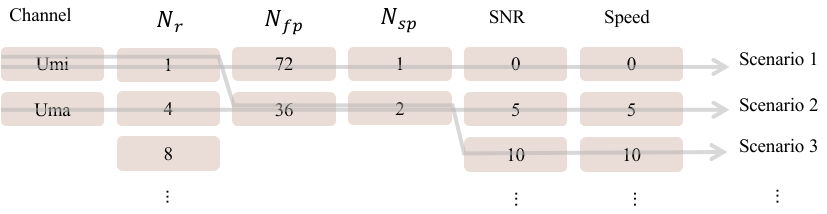}
    \caption{\textbf{Scenario structure}. We break down each scenario into six configurable parameters. }
    \label{fig:sc_struct}
\end{figure}
\noindent\textbf{Scenario Selection.}
Ideally, we would train and evaluate the models on all possible scenarios. However, the scenario's parameters themselves have expansive spaces. For this reason, we identify a selection of scenarios that cover different wireless communication settings as depicted in Fig.~\ref{fig:sc_struct}. We base our selection on three factors: (i) we prioritize scenarios that are more realistic than others, (ii) we take into account scenarios that are widely studied in the literature, and (iii) we design targeted scenarios to evaluate different metrics (e.g., accuracy, robustness, generalization). In the next section, we present the core\footnote{The use of the term \emph{core} does not suggest that any scenario in this set is more fundamental than ones outside the set.} scenarios considered in this benchmark.

\subsection{Core scenarios}
We design a set of core scenarios, illustrated in Table \ref{tab:task_summary}, to standardize the training and evaluation of deep channel estimators along different evaluation metrics.
the models along different evaluation metrics.
We are interested in three metrics: (i) the accuracy which evaluates the estimation performance, (ii) the domain generalization (a.k.a out-of-distribution (OOD) generalization)which investigates the performance under a distribution shift, and (iii) the robustness to spatial correlation and pilot design. The SNR values are sampled from an interval with a fixed stepsize of $5$. The UE speeds are also defined in the interval $[0,15]$ m/s with a stepsize of $5$ unless stated otherwise. For the channel model, we consider 3GPP 3D propagation models \footnote{\href{https://www.etsi.org/deliver/etsi_tr/138900_138999/138901/16.01.00_60/tr_138901v160100p.pdf}{3GPP TR 38.901}} (e.g 3D-Micro Urban (Umi) and 3D-Macro Urban (Uma)). As for the pilot design, we implemented single and double-symbol pilot patterns with different spacing in the frequency domain similar to the demodulation reference signal configurations in LTE and NR PUSCH transmissions \footnote{\href{https://www.etsi.org/deliver/etsi_ts/136200_136299/136211/14.02.00_60/ts_136211v140200p.pdf}{3GPP TS 36.211}, \href{https://www.etsi.org/deliver/etsi_ts/138200_138299/138211/15.02.00_60/ts_138211v150200p.pdf}{3GPP TS 38.211}}. 

 \subsection{Models}\label{sec:cebed_algos}
 Currently, CeBeb implements the following baselines \footnote{We did our best to follow the implementation details in the original papers although in some papers the network architectures are vaguely described.}: 
\begin{figure*}[ht!]
    \centering
    \begin{subfigure}[b]{0.5\textwidth}
    \centering
    \includegraphics[width=\textwidth]{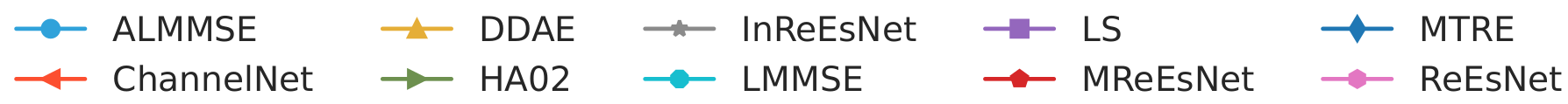}
     \end{subfigure}
    \begin{subfigure}[b]{\textwidth}
        \centering
        \includegraphics[width=0.33\textwidth]{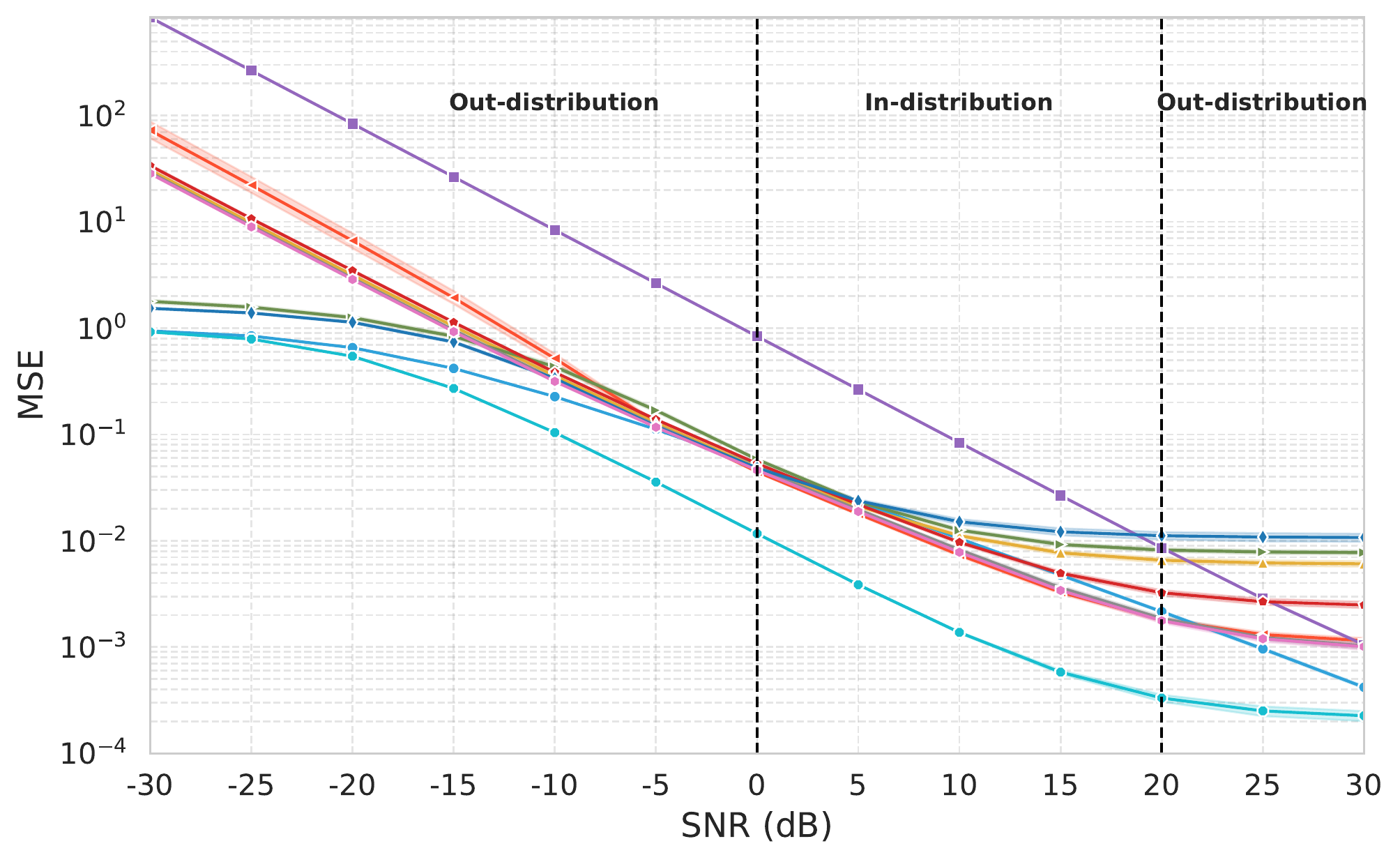}%
        \hfill
        \includegraphics[width=0.33\textwidth]{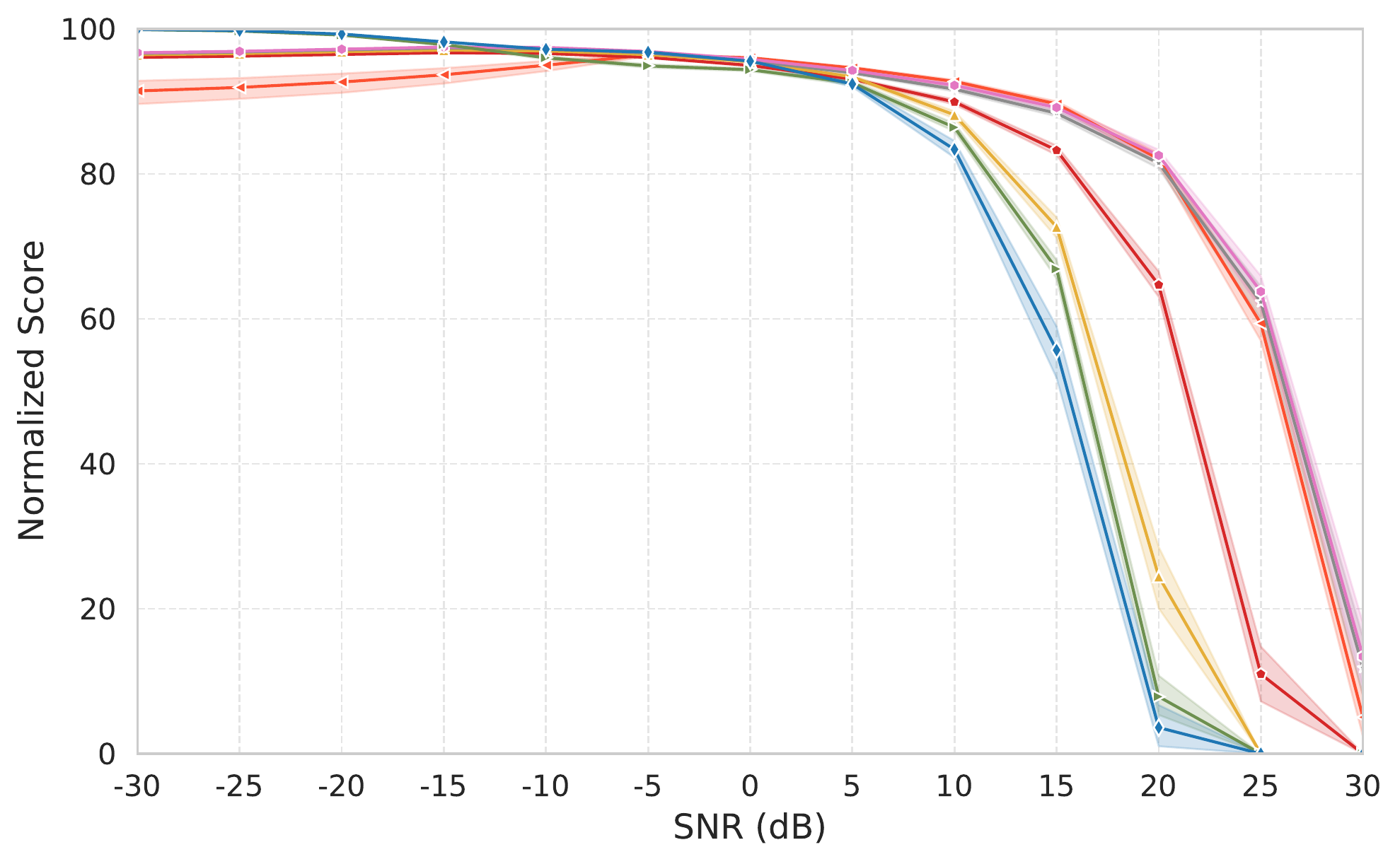}%
        \hfill
        \includegraphics[width=0.33\textwidth]{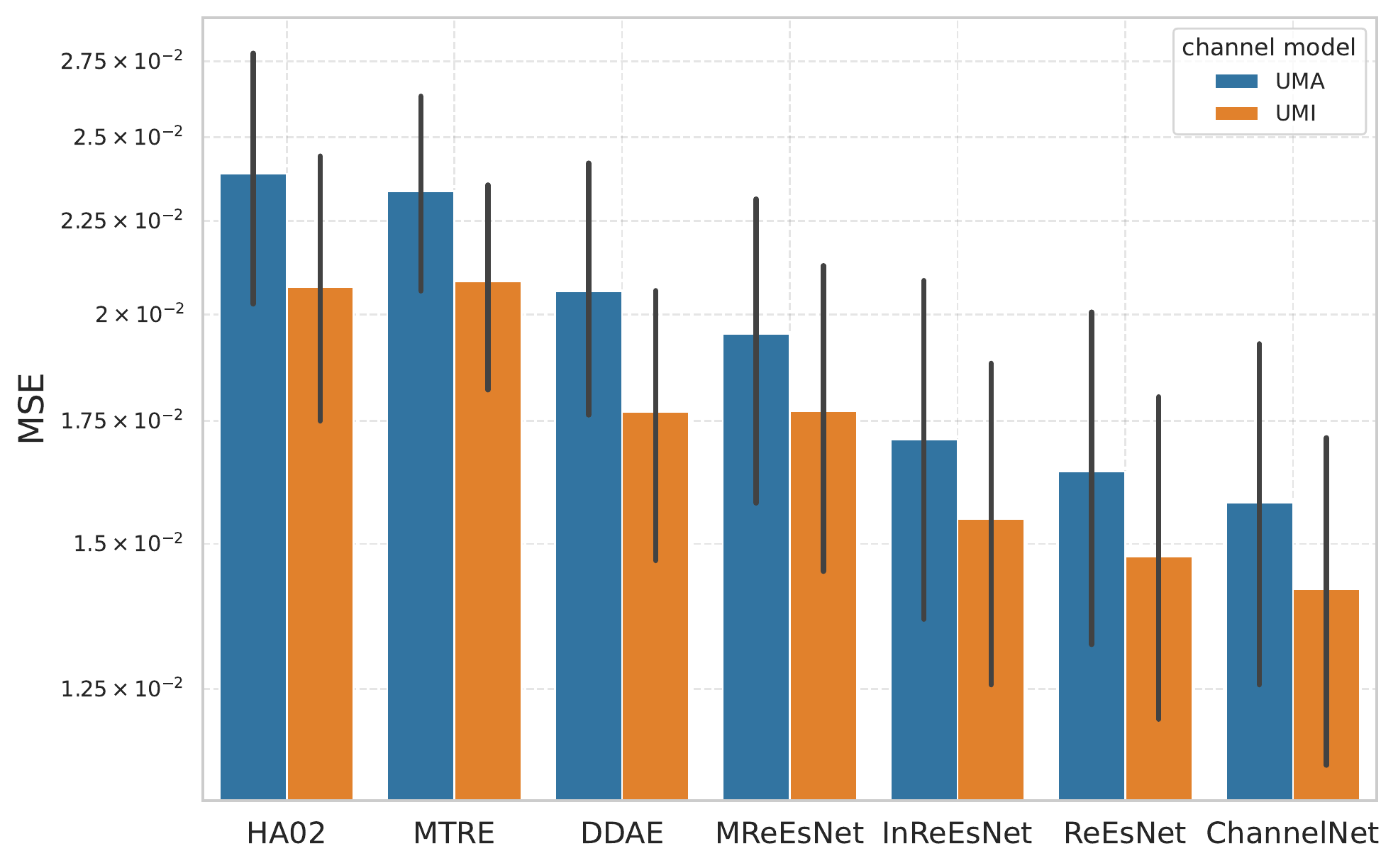}
        \caption{SISO case}\label{fig:siso_ind}
    \end{subfigure}
    \begin{subfigure}[b]{\textwidth}
        \centering
        \includegraphics[width=0.33\textwidth]{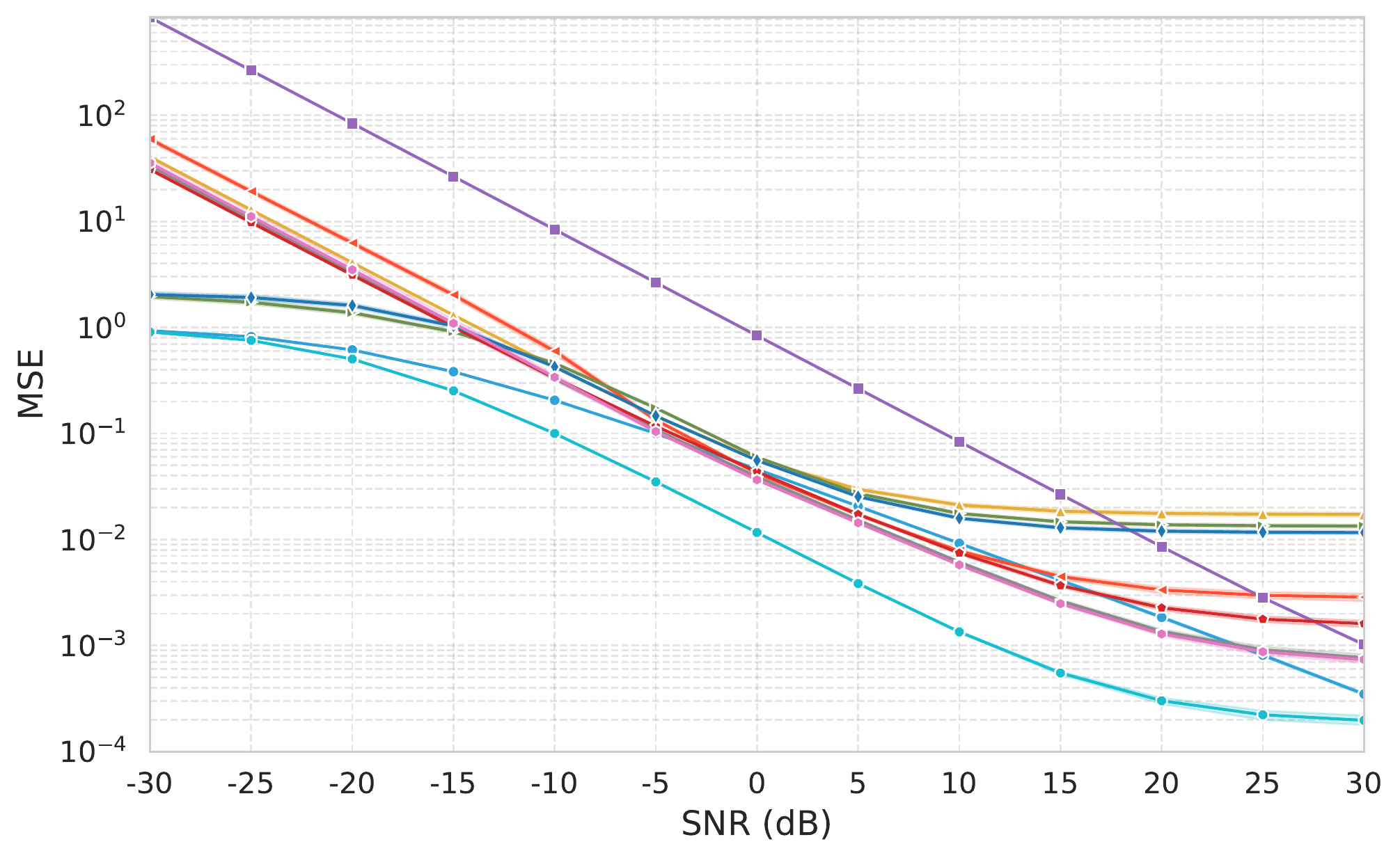}%
        \hfill
        \includegraphics[width=0.33\textwidth]{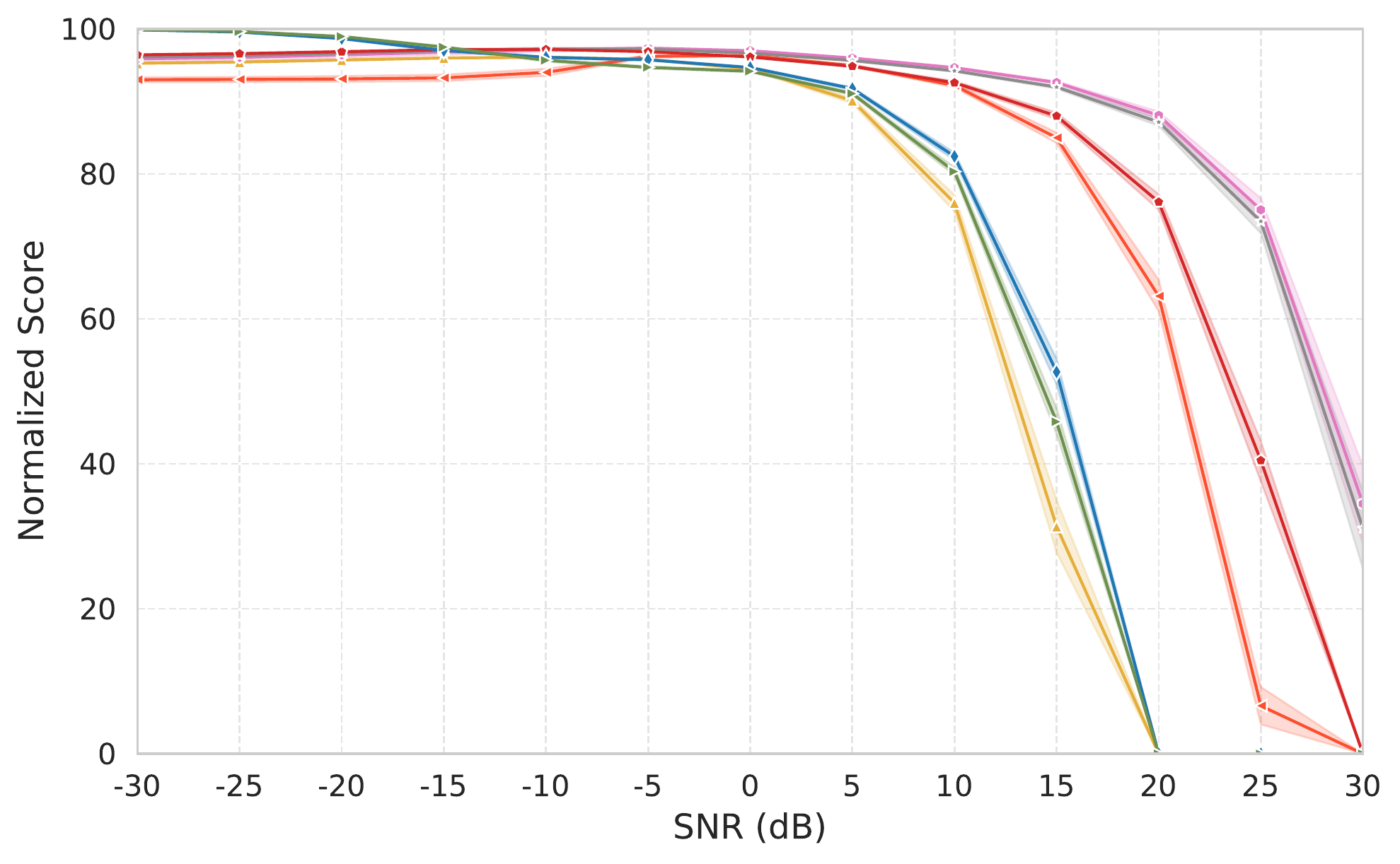}%
        \hfill
        \includegraphics[width=0.33\textwidth]{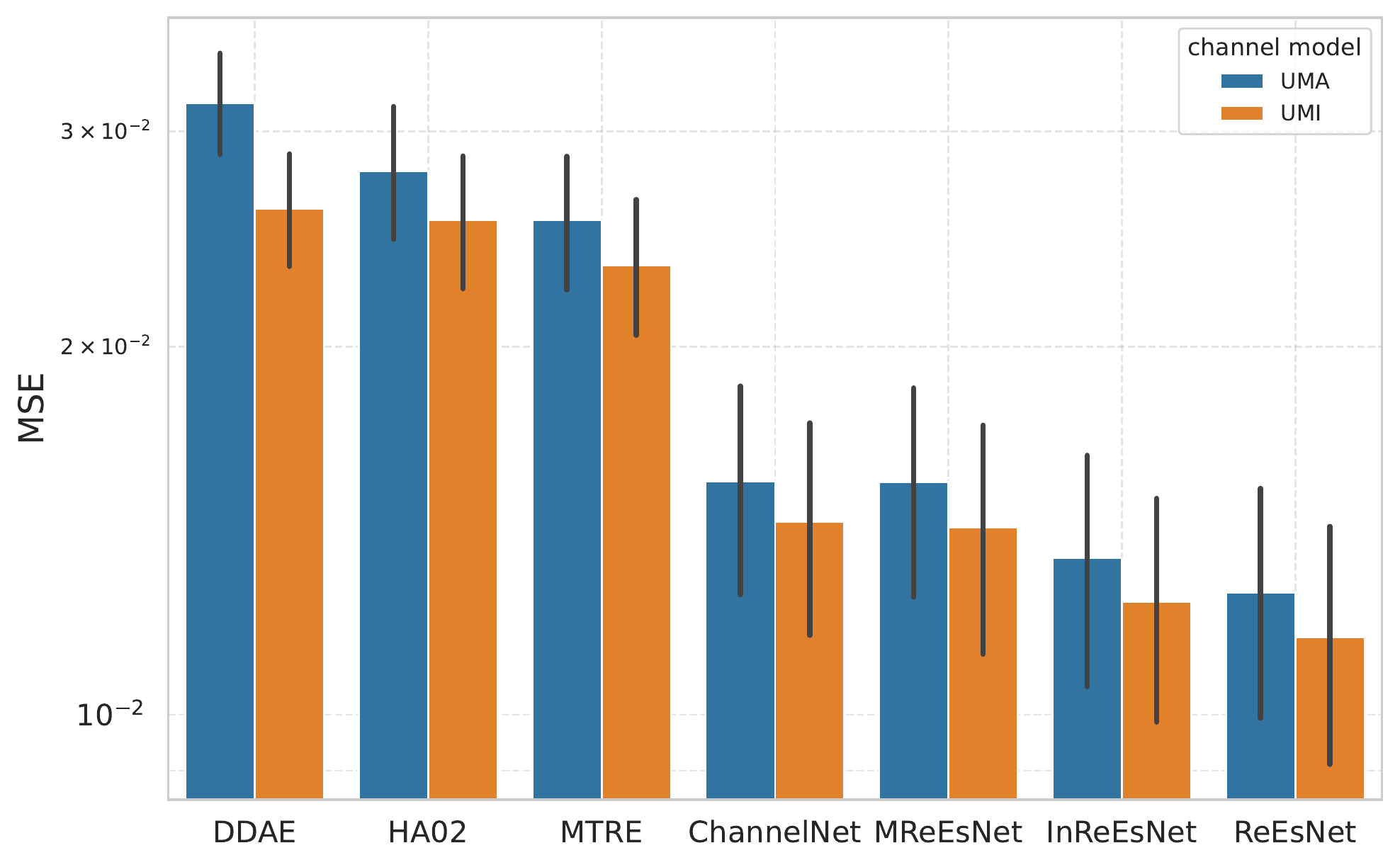}
        \caption{SIMO case}\label{fig:simo_ind}
    \end{subfigure}
    \caption{In-distribution performance: (\textbf{left}) The MSE as function of the SNR, (\textbf{center}) The normalized score, (\textbf{right}) The impact of the channel model.}\label{fig:avg_perf}

\end{figure*}
 \begin{itemize}[leftmargin=*]
     \item \textbf{ChannelNet} \cite{channelNet}: is a pre-sampling SR method that uses SRCNN to upscale the low-resolution inputs and a denoising CNN \cite{dccnn} to further refine the obtained channels. 
     Our implementation uses bilinear interpolation to upscale the low-resolution inputs and trains the SR and the denoiser in an end-to-end fashion; 
     \item \textbf{ReEsNet} \cite{ReEsNet} uses a post-sampling residual SR network (as in ESDR) and a deconvolution layer as an upscaling layer; 
     \item \textbf{InReEsNet} \cite{InReEsNet} is an extension of ReEsNet where the deconvolution layer is replaced by a bilinear interpolation;
     \item \textbf{MReEsNet} is another extension of ReEsNet, that we introduce, where the low-resolution inputs are replaced by the masked LS estimates;
     \item \textbf{DDAE} \cite{DDAE} is a dense denoising autoencoder. It takes the masked LS estimates as inputs and simultaneously denoises the coefficients at the pilot positions and estimates the values at the masked data locations;
     \item \textbf{MTRE} \cite{MTRE} is an MIM approach. It embeds the masked LS estimates using a 1D convolution layer, and then applies a sequence of Transformer encoder blocks; 
     \item \textbf{HA02} \cite{channelformer} is a hybrid auto-encoder that uses a Transformer encoder \cite{transformer} and a residual decoder (as in ReEsNet) and an up-sampling module. 
 \end{itemize}
\addtolength{\topmargin}{0.03in}
\section{Experimental Setup}\label{sec:res}
\noindent \textbf{Dataset Generation}: We use the open source link level simulator Sionna \cite{hoydis2022sionna}. For each scenario, we generate a dataset of $15000$ samples.
For all the datasets, we consider a propagation environment with path loss and non-line-of-sight. The carrier frequency and subcarrier spacing are $2.1$ GHz and $30$ kHz, respectively. The number symbols and subcarriers are $14$ and $72$. We consider block pilot design where pilot symbol locations are $3$ and $10$.
The values at pilot locations are random QPSK symbols.
All the channel realizations are normalized over the resource grid such that each resource element has a unit average energy.\\
\textbf{Training details}:
We randomly split all the datasets into $80\%$, $10\%$, $10\%$ splits. We use the large split for model training, one split for validation and one split for model evaluation. All the models are trained using Adam optimizer \cite{kingma2014adam} with an initial learning rate of $0.001$. We reduce the learning rate by half when the validation loss (i.e., MSE) stops improving for $3$ consecutive epochs. The minimum learning rate is $10^{-5}$. The training is stopped if the validation loss does not improve after $10$ consecutive epochs. 
For all experiments, we follow the literature (e.g., \cite{channelformer, ReEsNet}) and construct a multi-domain dataset using different SNR levels. The train SNR domains are between $0$ and $20$ (see ~\ref{tab:task_summary}). The models are trained on batches of $512$ samples randomly selected from each domain. \\
\noindent\textbf{Evaluation protocol}: The evaluation set contains $7500$ samples (i.e., ($\mathbf{X}, \mathbf{H}$) pairs). 
We repeat our experiments for each dataset \emph{five} times using five different random seeds which will result in different model initializations and dataset splits. The reported results are a mean and the $95\%$ confidence interval over these repetitions. We consider the baselines described in Section \ref{sec:background}: LS, LMMSE, and ALMMSE. The LS and LMMSE methods are widely used in the literature as upper and lower bounds, respectively, for the DNN performance.

In addition to mean square error (MSE), we introduce a new evaluation score called :
\begin{align}
    \textrm{Normalized\_score}=100\cdot\max\left(0,\frac{\textrm{MSE}_{\textrm{NN}}-\textrm{MSE}_{\textrm{LS}}}{\textrm{MSE}_{\textrm{LMMSE}}-\textrm{MSE}_{\textrm{LS}}}\right),\nonumber
\end{align}
where $\textrm{MSE}_{\textrm{LS}}, \textrm{MSE}_\textrm{LMMSE}$ and $\textrm{MSE}_\textrm{NN}$ denote the MSE of the LS, LMMSE and NN-based methods, respectively.
A normalized score close to $100\%$ means that the model brings the LS performance closer to the LMMSE one. Whereas, a normalized score equal to $0$ indicates a failure mode where the model is not able to enhance the LS performance.
 \section{Results}

\begin{table}[ht!]
    \centering
    \caption{Evaluation of the accuracy \& generalization}
    
    \label{tab:mse_gain}
    \begin{tabular}{l| l l | l l}
        Method & \multicolumn{2}{l}{\thead{In-Distribution \\ Train scenarios \\ = \\Test scenarios}} & \multicolumn{2}{l}{\thead{OOD \\Train scenarios \\ $\neq$ \\Test scenarios}} \\
        \cline{2-5}
        & MSE & Gain (dB) & MSE & Gain (dB)\\ 
        \hline
        LS & $0.24$ & - & $152.81$ & - \\
        ALMMSE & $0.017$& $11.47$ & $0.39$ & $25.92$\\
        LMMSE & $0.004$ & $18.37$ & $0.33$ & $26.71$ \\
        \hline
        ChannelNet & $0.015$ & $12.13$ & $11.97$ & $11.06$\\
        ReEsNet & $\textbf{0.014}$& $\textbf{12.47}$ & $5.83 $ & $14.19$\\
        InReEsNet & $0.015$& $12.24$ & $5.70$ & $14.28$\\
        MReEsNet & $0.017$& $11.64$ & $5.92$ & $14.12$\\
        DDAE & $0.024$& $10.08$ & $6.54$ & $13.96$\\
        MTRE & $0.023$& $10.22$ & $\textbf{0.78}$ &$\textbf{22.92}$ \\
        HA02 & $0.024$& $9.99$ & $0.79$ & $22.85$\\
        \bottomrule
        
    \end{tabular}
    
\end{table}
\subsection{Accuracy: In-distribution Performance}
To evaluate the accuracy, we train models for each train scenario and compute their performance on the same scenarios. Recall that each scenario has separate train and test sets. This setting is called in-distribution since the train and test scenarios are the same, thus the data distribution doesn't change. For each model, we report the average MSE across all scenario parameters and seeds. 
We start by analyzing the average performance of all models reported in Table \ref{tab:mse_gain}. For the in-distribution scenario, ReEsNet and InReEsNet achieve the best average performance compared to all deep baselines and ALMMSE with $12.5$ and $12.2$ dB gain, respectively, compared to LS. We present the detailed performance of these models as a function of the SNR in Fig. ~\ref{fig:avg_perf}. For low SNR regime ($\rho=1$ or $5$ dB), the performance of all deep models is close. However, in a high SNR regime, a pattern emerges where the performance stagnates and even becomes worse than LS for MTRE, HA02, and DDAE. Amongst the MIM methods (i.e., MTRE, MResNet, and DDAE), MResNet performs best with a normalized score above $80 \%$. As for the transformer-based approaches, HA02 showcases better performance than MTRE in the SISO case which is not the case for SIMO. Next, we will analyze the impact of the system and channel models on performance. It is clear from Figs.~\ref{fig:siso_ind} and ~\ref{fig:simo_ind} that the performance of some models changes when the system model changes (SISO vs. SIMO). As an example, ChannelNet (HA02) outperforms MResNet (MTRE) in the SISO case, but this is reversed in the SIMO case. Furthermore, we observe that all models perform slightly better when the UMI channel model is used.
\subsection{Domain Generalization: OOD Performance}
Here, we analyze how the models' performance behaves when the models are evaluated on different SNR levels other than the ones used in training. The SNR levels are in $[0,20]$ and $[-30,0[ \cup [25,30]$ for the train and test domains, respectively. We focus on the domain generalization to changes in the SNR level since previous art reported that change in the user speed (i.e., doppler shift) does not harm the performance \cite{channelformer}. Also, it was shown that changes in SNR levels significantly alter the distribution of the received signals \cite{akrout2022continual}. 

We report in Table ~\ref{tab:mse_gain} the average MSE performance for the out-distribution case. Note that the results are averaged over the test domains only ($[-30,0[ \cup [25,30]$). 
Transformer-based approaches have the best performance in low SNR regimes ($[-30,0[$), in fact they have a close performance to LMMSE and hugely improve on the LS estimator. However, the performance of these models saturates and becomes worse than LS at a high SNR regime ($\geq 25$). On average, MTRE and HA02 achieve a gain close to $23$ dB when compared to LS, $62\%$ better than ReEsNet. This experiment shows that transformer-based models work best for denoising the LS estimates. This is consistent with other findings in the literature showing that the self-attention mechanism enables better robustness to noise corruptions than CNN-based architectures \cite{bai2021are}. Although HA02 is a hybrid architecture, the usage of a transformer encoder helped improve drastically the robustness of the ReEsNet-like architecture. This robustness comes with a cost since it leads to a saturation and the model stops improving as the noise level decreases. Further investigations are needed to understand why the transformer-based approaches fail in high SNR regimes. 
\subsection{Robustness to Pilot Design}
\begin{figure}[!ht]
    \centering
    \includegraphics[scale=0.4]{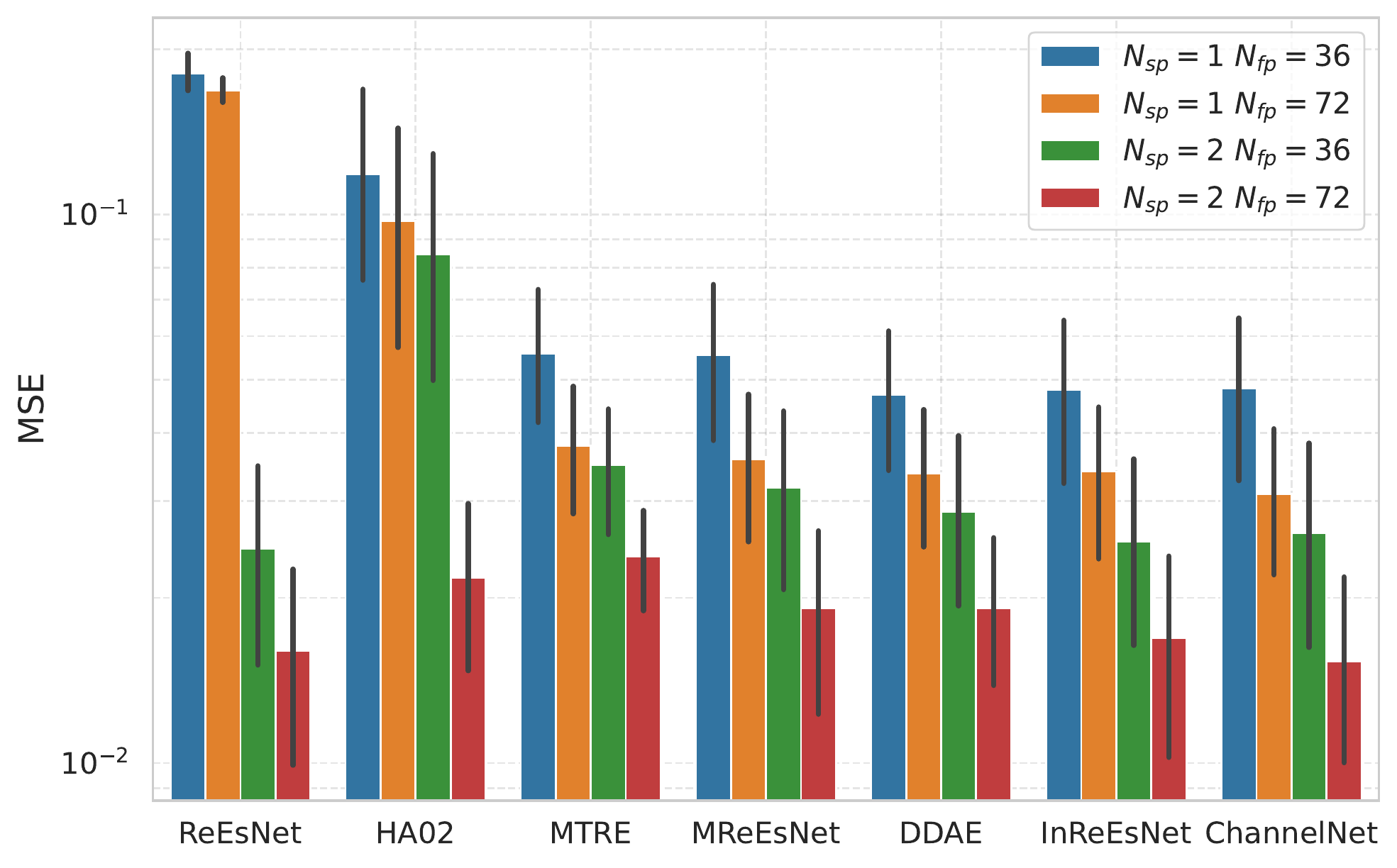}
    \caption{Average performance for different pilot designs.}
    \label{fig:pilot}
\end{figure}

In this section, we study the impact of the pilot design on the performance. For this analysis, we fix the other system parameters and only modify the pilot patterns. Unlike previous works, we analyze the impact of pilot arrangements in time and frequency instead of the total number of pilots (e.g., \cite{ReEsNet}). 
As depicted in Fig~\ref{fig:pilot}, the performance of all the models deteriorates as the number of pilots decreases. Particularly, ReEsNet's performance is hugely impacted when only one pilot symbol is used. Indeed, although the pilot arrangements ($N_{fp}=72, N_{sp}=1$) and ($N_{fp}=36, N_{sp}=2$) have the same number of pilots, the ReEsNet performance is drastically lower when $N_{sp}=1$. ChannelNet showcases the best robustness. Interestingly, InReEsNet has superior performance than ReEsNet despite the small change in the upscaling layer.
\subsection{Robustness to Special Correlation}
\begin{figure}[!ht]
    \centering
    \includegraphics[scale=0.4]{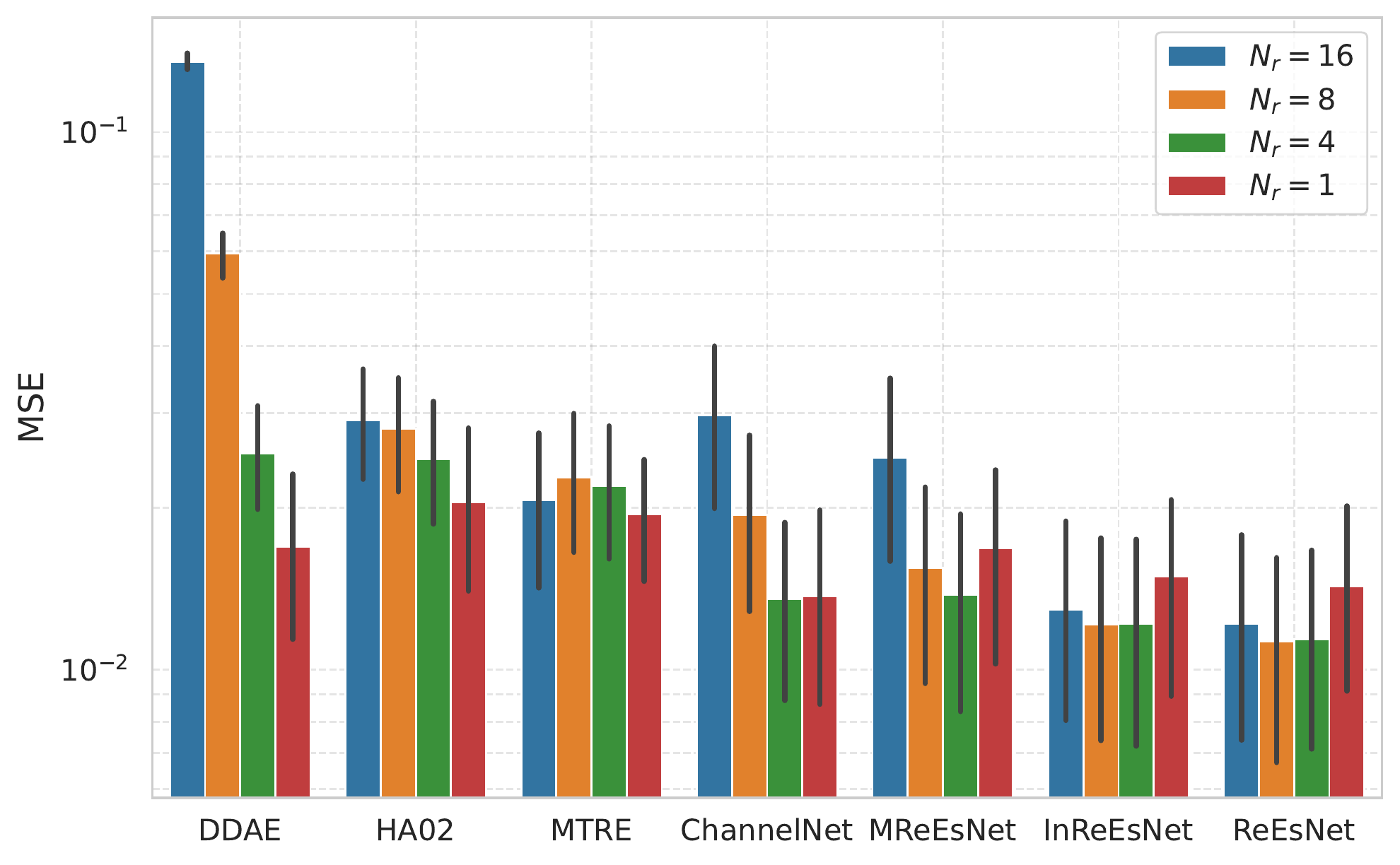}
    \caption{Average performance for different $N_r$ values.}
    \label{fig:nr}
\end{figure}

Our work is the first effort to investigate the impact of spatial correlation on the performance of deep models.
Figure ~\ref{fig:nr} reports the MSE as a function of the number of received antennas. Again, for this experiment, we fix all the scenario parameters and only vary $N_r$.
We observe that some models are robust to spatial correlation including ReEsNet, InReEsNet and MTRE. For ReEsNet and InReEsNet, the performance improves when $N_r$ increases. This means that these architectures can learn how to capture the spatial correlation at the receiver. Surprisingly, this is not the case for MReEsNet, which uses the same architecture as ReEsNet but with the masked LS estimates instead of the low-resolution inputs. Additionally, MTRE is the only model where the performance is nearly the same for different $N_r$ values.
This can be justified by the self-attention mechanism in MTRE which is conducted in frequency and space domains. This is another intriguing property of the self-attention mechanism that calls for further investigations. For the rest of the models (DDAE, HA02, ChanneNet), the performance deteriorates as $N_r$ increases. 
\section{Lessons Learned \& Conclusion}
We proposed CeBed, an open-source benchmark for deep channel estimation. The benchmark is designed to analyze different performance metrics under a standardized experimental setup. In addition to our experimental findings, we share lessons learned during the journey of building CeBed: 
\begin{itemize}[leftmargin=*]
    \item \textbf{Reproducible research.} Previous works only share the dataset without the generation process or a proper documentation. More recent endeavors (e.g.,\cite{belgiovine2021}) are publishing implementations to generate datasets for channel estimation using MATLAB, only accessible to licensed MATLAB users. In addition, we notice a serious lack of information regarding the architecture and the training which makes reproducing the results impossible;
    \item \textbf{Towards wireless-based solutions}: the baselines in this work are applying common computer vision architectures with minor modifications. To design a robust and accurate deep channel estimator, we need to go beyond simple applications and incorporate more wireless knowledge in the design of these solutions;
    \item \textbf{Beyond supervised learning}: One limitation of the baseline models is they are trained in a supervised manner and assume perfect knowledge of the true channels. This assumption is unrealistic in more practical implementations.
\end{itemize}
We aspire that CeBed can help the community by providing a unified framework for training and evaluation. In future work, we aim to continue building towards a more complete benchmark including more scenarios and models.

\bibliographystyle{IEEEtran}

\end{document}